\documentclass[lettersize,journal]{IEEEtran} 
\usepackage{amsmath,amsfonts}
\usepackage{algorithm}
\usepackage{algpseudocode}
\usepackage{array}
\usepackage[caption=false,font=normalsize,labelfont=sf,textfont=sf]{subfig}
\usepackage{textcomp}
\usepackage{stfloats}
\usepackage{url}
\usepackage{verbatim}
\usepackage{graphicx}
\usepackage{cite}
\usepackage{amsmath}
\usepackage{amssymb}
\usepackage{xcolor}
\usepackage{pifont}
\usepackage{booktabs}
\usepackage{multirow}
\usepackage{bbding}
\usepackage{orcidlink}
\usepackage{bbm}
\usepackage{colortbl}
\usepackage[capitalise]{cleveref}
\begin{document}

\title{\emph{FoQuS}: A Forgetting-Quality Coreset Selection Framework for Automatic Modulation Recognition}

\author{Yao Lu\orcidlink{0000-0003-0655-7814}, ~\IEEEmembership{Student Member, IEEE}, Chunfeng Sun, Dongwei Xu\orcidlink{0000-0003-2693-922X},~\IEEEmembership{Member, IEEE}, Yun Lin\orcidlink{0000-0003-1379-9301},~\IEEEmembership{Senior Member, IEEE}, Qi Xuan\orcidlink{0000-0002-6320-7012},~\IEEEmembership{Senior Member,~IEEE}, Guan Gui,~\IEEEmembership{Fellow,~IEEE}
\thanks{This work was partially supported by the National Natural Science Foundation of China under Grant 62301492, 61973273, U21B2001 and by the Key R\&D Program of Zhejiang under Grant 2022C01018. (Corresponding author: Qi Xuan, Dongwei Xu)}
\thanks{Yao Lu, Chunfeng Sun, Dongwei Xu and Qi Xuan are with the Institute of Cyberspace Security, College of Information Engineering, Zhejiang University of Technology, Hangzhou 310023, China, also with the Binjiang Institute of Artificial Intelligence, Zhejiang University of Technology, Hangzhou 310056, China (e-mail: yaolu.zjut@gmail.com, 221124030344@zjut.edu.cn, dongweixu@zjut.edu.cn, xuanqi@zjut.edu.cn).}
\thanks{Yun Lin is with the College of Information and Communication Engineering, Harbin Engineering University, Harbin, China (e-mail: linyun@hrbeu.edu.cn).}
\thanks{Guan Gui is with the College of Telecommunications and Information Engineering, Nanjing University of Posts and Telecommunications, Nanjing 210003, China (e-mail: guiguan@njupt.edu.cn).}}

\markboth{Journal of \LaTeX\ Class Files,~Vol.~14, No.~8, August~2021}%
{Shell \MakeLowercase{\textit{et al.}}: A Sample Article Using IEEEtran.cls for IEEE Journals}


\maketitle

\begin{abstract}
Deep learning-based Automatic Modulation Recognition (AMR) model has made significant progress with the support of large-scale labeled data. However, when developing new models or performing hyperparameter tuning, the time and energy consumption associated with repeated training using massive amounts of data are often unbearable. To address the above challenges, we propose \emph{FoQuS}, which approximates the effect of full training by selecting a coreset from the original dataset, thereby significantly reducing training overhead. Specifically, \emph{FoQuS} records the prediction trajectory of each sample during full-dataset training and constructs three importance metrics based on training dynamics. Experiments show that \emph{FoQuS} can maintain high recognition accuracy and good cross-architecture generalization on multiple AMR datasets using only 1\%-30\% of the original data.
\end{abstract}


\section{Introduction}
\IEEEPARstart{I}{n} recent years, deep learning-based automatic modulation recognition (AMR) technology has made significant progress~\cite{wang2019data,lin2020adversarial}. Unlike traditional expert-based feature extraction methods, deep learning models learn complex feature representations directly from raw signal data, enabling higher recognition accuracy in a variety of complex scenarios. However, training these deep learning models usually relies on large-scale labeled datasets~\cite{o2016radio,o2018over,chen2021signet} and a lot of computing resources. In many practical scenarios, such overhead is not affordable. For example, when designing a new model structure or adjusting hyperparameters, researchers often need to quickly verify its performance on a certain task. Using the full training set for validation may take hours to days (depending on the data size and model capacity), which not only significantly slows down the experimental cycle but also consumes a lot of resources. To alleviate these issues, we consider whether it is possible to select a smaller but representative subset from the full dataset and use it instead of the original dataset for training. 

Coreset selection~\cite{sener2017active,margatina2021active,paul2021deep,toneva2018empirical} is a method proposed to address this need. It aims to approximate the performance of the full training set with a smaller sample set, thereby achieving efficient model development. Although existing coreset methods have achieved remarkable results on image tasks, these methods can not be directly transferred to AMR tasks due to the essential differences in data structure between signals and images. Furthermore, signal datasets have a crucial physical property: signal-to-noise ratio (SNR). Ignoring SNR during selection may lead to oversampling of high-SNR, easily learned samples, while overlooking low-SNR samples. This will lead to significant performance degradation in noisy environments.

To this end, we propose a \textbf{Fo}rgetting–\textbf{Qu}ality \textbf{S}core (\emph{FoQuS}) for AMR tasks. Specifically, \emph{FoQuS} first trains a model on the entire dataset and records the prediction trajectory of each example. It then constructs three complementary metrics: \textbf{Forgetting Score}, \textbf{Persistent Error Score}, \textbf{Quality Score}. The forgetting score aims to locate samples close to the decision boundary that are easily "forgotten" by parameter perturbations; the persistent error score captures stubborn examples that are difficult to fit during training; and the quality score focuses on samples that are partially learned by the model but can still provide meaningful gradients. We normalize the three metrics and add them together to derive the final \emph{FoQuS} score. Instead of simply selecting the global top-k samples based on \emph{FoQuS} scores, we first divide the samples into three levels based on the scores and then extract a corresponding proportion of samples from each level according to a preset ratio to form a diversified coreset. In subsequent experiments, we show that the coresets obtained by \emph{FoQuS} outperform existing image-based coreset selection methods on multiple AMR tasks. To summarize, our main contributions are three-fold:


\begin{itemize}
    \item We propose \emph{FoQuS} to filter a small and precise subset from the original signal dataset for subsequent model development.
    \item Our \emph{FoQuS} score consists of a normalized combination of a forgetting score, a persistent error score and a quality score, which together capture complementary aspects of the importance of each sample.
    \item We conduct extensive experiments on $3$ AMR datasets, demonstrating that our method outperforms $10$ coreset selection baselines and exhibits better cross-architecture generalization capabilities.
\end{itemize}

In the remainder of this paper, we first introduce related works on coreset selection in \cref{sec:Related Works}. In \cref{sec:model purify}, we delve into our coreset selection method. Then experiments are discussed in \cref{sec:Experiments}. Finally, the paper concludes in \cref{sec:Conclusion}.

\section{Related Works}
\label{sec:Related Works}

\subsection{Coreset Selection} 
Coreset selection aims to construct a compact yet effective subset of training examples that maintains the original model’s performance while reduces the training cost. It typically design a metric to measure the importance of each sample in the training set to the entire dataset, and selects a coreset based on this metric to replace the entire original training set. Coreset selection can be applied to a wide range of downstream scenarios, including active learning~\cite{sener2017active}, training acceleration~\cite{xia2022moderate}, data distillation~\cite{tukan2023dataset}, continual learning~\cite{yoon2021online} and transfer learning~\cite{zhang2025non}. In the past ten years, researchers have proposed a variety of methods. For example, Forgetting~\cite{toneva2018empirical} counts the number of times a sample changes from a correct prediction to an incorrect prediction during training, giving priority to retaining samples that are easily forgotten. 
To advance the research of coreset selection, Guo et al.~\cite{guo2022deepcore} contribute a comprehensive code library, namely DeepCore. Recently, Lu et al.~\cite{lu2024rk} introduce RK-core to understand the intricate hierarchical structure within datasets. Besides, some studies have extended coreset selection to tasks such as object detection~\cite{lee2024coreset} and federated learning~\cite{hao2025fedcs}.

Although research on coreset selection has made great progress in recent years, they mainly focus on the image field. In contrast, coreset selection has not been explored in the field of wireless signal processing and AMR. We believe that this phenomenon is caused by: 1) the structured nature of signal data in the time-frequency domain makes direct image-based methods ineffective; 2) image tasks usually only need to focus on class balance, while the signal also carries the key signal-to-noise ratio (SNR) characteristics. If the SNR factor is not considered during selection, directly transplanting image-based methods is likely to lose certained SNR samples, resulting in a significant drop in performance under noisy conditions. In light of these challenges, this paper proposes an effective coreset selection strategy for AMR tasks, filling a gap in its application in signal processing.

\section{Method}
\label{sec:model purify}
In this section, we first give the mathematical definition of core set selection and then elaborate on our proposed method. 
\subsection{Problem Definition}
\label{sec:problem definition}
In this subsection, we provide a mathematical definition of the coreset selection problem. Specifically, given a training set $\mathcal{D}=\{(x_i,y_i)\}_{i=1}^N$, coreset selection aims to construct a compact yet effective subset $\mathcal{S}=\{(x_i,y_i)\}_{i=1}^{|\mathcal{S}|}$, where $N \gg |\mathcal{S}|$, so that the model trained on this subset can perform similarly to the model trained on $\mathcal{D}$. Mathematically, we can define the problem as follows:
\begin{equation}
\label{eq:loss_full}
\begin{aligned}
    \min _{\mathcal{S} \subseteq \mathcal{D},|\mathcal{S}| \leq k}&\left|\mathcal{L}_{\mathcal{D}}\left(\theta_{\mathcal{D}}\right)-\mathcal{L}_\mathcal{S}\left(\theta_\mathcal{S}\right)\right|,\\
    \textit{s.t.,} \quad\mathcal{L}_{\mathcal{D}}(\theta_{\mathcal{D}})=&\frac{1}{N}\sum_{(x,y)\in\mathcal{D}}\ell(\theta; x,y),\\
    \mathcal{L}_{\mathcal{S}}(\theta_{\mathcal{S}})=&\frac{1}{|\mathcal{S}|}\sum_{(x,y)\in \mathcal{S}}\ell(\theta; x,y).\\
\end{aligned}
\end{equation}
where $\theta_{\mathcal{D}}$ and $\theta_{\mathcal{S}}$ denote the model parameter learned from the full dataset $\mathcal{D}$ and the selected coreset $\mathcal{S}$, respectively. $\ell$ denotes the loss function used for training the model. $k$ is the data sampling rate and is calculated by $k = \frac{|\mathcal{S}|}{N}$.

\subsection{\emph{FoQuS} Score}
To obtain a coreset with excellent performance, the selected subset must be selected to cover the characteristic distribution of the original dataset. In AMR tasks, this means that the coreset must cover both different modulation categories and different SNR ranges. However, traditional image-based coreset selection methods~\cite{margatina2021active,paul2021deep,toneva2018empirical} often select samples based on only a single metric, which can easily lead to insufficient coverage, resulting in poor performance of the trained model in noisy settings or specific scenarios. To this end, we advocate using \emph{diversity metrics} to evaluate the importance of each sample. These metrics can reflect the contribution of the sample to the original dataset from multiple perspectives in a complementary manner. In this paper, our \emph{FoQuS} score consists of the following three parts:
\begin{itemize}
    \item \textbf{Forgetting Score} reveals instabilities in the training process and can locate samples near the decision boundary that are easily “forgotten” by parameter perturbations.
    \item \textbf{Persistent Error Score} is used to capture difficult samples that cannot be fit for a long time.
    \item \textbf{Quality Score} measures the learning value of each sample, focusing on simple, easily classified samples.
\end{itemize}
We provide a detailed description of these scores below. 


Firstly, we train a model \(f_{\theta}\) on the full training set \(\mathcal{D}=\{(x_i,y_i)\}_{i=1}^N\) for \(T\) epochs using standard stochastic gradient descent algorithm. During training we can obtain the predicted label for each sample and record it.
\begin{equation}
\label{eq:predicted label}
\hat y_i^{(t)}= f_{\theta}^{(t)}(x_i), \quad t=1, \ldots, T.
\end{equation}
Then we can calculate the correctness of each sample based on the predicted label and the correct label as follows:
\begin{equation}
\label{eq:correctness}
c_i^{(t)}=\mathbf{1}\left[\hat{y}_i^{(t)}=y_i\right]
\end{equation}
According to the correctness, we can further obtain the forgetting score $S_{\textit{forget}}$ and persistent error score $S_{\textit{persist\_err}}$ of each sample.
\begin{equation}
\label{eq:forget}
S_{\textit{forget}}:=\sum_{t=2}^T 1\left[c_i^{(t-1)}=1 \wedge c_i^{(t)}=0\right],
\end{equation}
\begin{equation}
\label{eq:per}
S_{\textit{persist\_err}}:=\sum_{t=2}^T 1\left[c_i^{(t-1)}=0 \wedge c_i^{(t)}=0\right] .
\end{equation}
At the same time, we accumulate the per-sample loss $\mathcal{L}_{\text{accum},i}$ and correctness counts $\mathcal{L}_{\text{count},i}$ in each epoch:
\begin{equation}
\label{eq:accum}
\mathcal{L}_{\text{accum},i}=\sum_{t=1}^T \text{CE}(\hat y_i^{(t)},y_i), \quad \mathcal{L}_{\text{count},i}=\sum_{t=1}^T c_i^{(t)},
\end{equation}
where $\text{CE}(\cdot)$ is the cross-entropy loss and $i$ denotes the $i$-th sample in the training dataset. Then the quality score can be calculated as follows:
\begin{equation}
\label{eq:quality}
S_{\textit{quality}}:=\frac{\mathcal{L}_{\text{count},i}}{T}-\beta \times \frac{\mathcal{L}_{\text{accum},i}}{T},
\end{equation}
where $\beta$ is a scaling factor and we set it to $0.1$ by default. On this basis, our \emph{FoQuS} score is formulated as:
\begin{equation}
\label{eq:FoQuS}
S_{\textit{FoQuS}}=\frac{S_{\textit{forget}}}{T-1}+\frac{S_{\textit{persist\_err}}}{T-1}+S_{\textit{quality}}.
\end{equation}
After obtaining the \emph{FoQuS} score of each sample, we do not simply take the top-$k$ scoring samples. Instead, we partition the dataset into three tiers according to the \emph{FoQuS} score and draw a specified proportion of samples from each tier to form the coreset. Finally, we summarize the overall pipeline of our approach in \cref{algorithm:framework}.


\begin{algorithm}[t]
    \caption{\emph{FoQuS} Algorithm}
    \label{algorithm:framework}
    \textbf{Input}: A training dataset $\mathcal{D} \in \{(x_i, y_i)\}_{i=1}^{N}$, training epoch $T$, a deep learning model $f_\theta$.\\
    \textbf{Output}: A coreset $\mathcal{S}$. \\
    \begin{algorithmic}[1] 
    \For{$t = 1$ to $T$}
        \State Train $f_\theta$ on $\mathcal{D}$ using stochastic gradient descent. 
        \State Obtain the predicted label via Eq.~\eqref{eq:predicted label}.
        \State Calculate the correctness of each sample using \cref{eq:correctness}.
        \State Calculate $S_{\textit{forget}}$ and $S_{\textit{persist\_err}}$ via \cref{eq:forget} and \cref{eq:per}.
        \State Accumulate the per-sample loss $\mathcal{L}_{\text{accum},i}$ and correctness counts $\mathcal{L}_{\text{count},i}$ using \cref{eq:accum}. 
        \State Calculate $S_{\textit{quality}}$ using \cref{eq:quality}.
        \State Obtain the final \emph{FoQuS} score using \cref{eq:FoQuS}.
    \EndFor
        \State Sort the \emph{FoQuS} score in descending order.
        \State Partition $\mathcal{D}$ into three tiers.
        \State Draw a proportion of samples from each tier to form $\mathcal{S}$.
    \end{algorithmic}
\end{algorithm}

\begin{table*}[t]
  \centering
  \caption{Comparison of our method against coreset selection baselines under various rates on RML2016.10a and Sig2019-12. The best results are marked in \textbf{boldface}. For each experimental result, we run $3$ times and report its mean.}
  \resizebox{0.99\textwidth}{!}{
    \begin{tabular}{cccccccc|cccccccc}
    \toprule
    \multicolumn{8}{c|}{RML2016.10a}                              & \multicolumn{8}{c}{Sig2019-12} \\
    \midrule
    \multirow{2}[4]{*}{Model} & \multirow{2}[4]{*}{Method} & \multicolumn{5}{c}{Rate}              & \multirow{2}[4]{*}{Avg} & \multirow{2}[4]{*}{Model} & \multirow{2}[4]{*}{Method} & \multicolumn{5}{c}{Rate}              & \multirow{2}[4]{*}{Avg} \\
\cmidrule{3-7}\cmidrule{11-15}          &       & 1\%   & 5\%   & 10\%  & 20\%  & 30\%  &       &       &       & 1\%   & 5\%   & 10\%  & 20\%  & 30\%  &  \\
    \midrule
    \multirow{11}[2]{*}{CNN1D} & Uniform & 0.2907  & 0.4587  & 0.5257  & 0.5572  & 0.5793  & 0.4823  & \multirow{11}[2]{*}{CNN1D} & Uniform & 0.5598  & 0.7314  & 0.7685  & 0.9034  & 0.9259  & 0.7778  \\
          & Cal   & 0.3782  & 0.5769  & 0.7319  & 0.8162  & 0.8258  & 0.6658  &       & Cal   & 0.4868  & 0.6193  & 0.7613  & 0.8786  & 0.9108  & 0.7314  \\
          & CD    & 0.4420  & 0.6857  & 0.7580  & 0.8263  & 0.8369  & 0.7098  &       & CD    & 0.4431  & 0.6485  & 0.7121  & 0.8909  & 0.9247  & 0.7239  \\
          & Forgetting & 0.3564  & 0.5873  & 0.7181  & 0.7867  & 0.8170  & 0.6531  &       & Forgetting & 0.4974  & 0.5280  & 0.6626  & 0.8594  & 0.8791  & 0.6853  \\
          & GraNd & 0.3934  & 0.7056  & 0.7678  & 0.8177  & 0.8409  & 0.7051  &       & GraNd & 0.4843  & 0.5579  & 0.7011  & 0.8380  & 0.8712  & 0.6905  \\
          & Herding & 0.4245  & 0.6579  & 0.7719  & 0.8156  & 0.8222  & 0.6984  &       & Herding & 0.5909  & 0.7360  & 0.7629  & 0.9129  & 0.9241  & 0.7854  \\
          & k-Center Greedy & 0.3801  & 0.6441  & 0.7827  & 0.8218  & 0.8459  & 0.6949  &       & k-Center Greedy & 0.5406  & 0.7341  & 0.7608  & 0.9127  & 0.9253  & 0.7747  \\
          & Least Confidence & 0.3077  & 0.6171  & 0.7182  & 0.7617  & 0.7930  & 0.6395  &       & Least Confidence & 0.4802  & 0.6671  & 0.7438  & 0.8657  & 0.8958  & 0.7305  \\
          & Entropy & 0.3993  & 0.5670  & 0.6288  & 0.7818  & 0.7960  & 0.6346  &       & Entropy & 0.4500  & 0.5720  & 0.7453  & 0.8500  & 0.8704  & 0.6975  \\
          & Margin & 0.4418  & 0.5962  & 0.6863  & 0.7576  & 0.8044  & 0.6573  &       & Margin & 0.4623  & 0.5955  & 0.6980  & 0.8610  & 0.8965  & 0.7027  \\
          & \emph{FoQuS}  & \cellcolor[rgb]{ .816,  .808,  .808}\textbf{0.5410} & \cellcolor[rgb]{ .816,  .808,  .808}\textbf{0.7066} & \cellcolor[rgb]{ .816,  .808,  .808}\textbf{0.8036} & \cellcolor[rgb]{ .816,  .808,  .808}\textbf{0.8282} & \cellcolor[rgb]{ .816,  .808,  .808}\textbf{0.8487} & \cellcolor[rgb]{ .816,  .808,  .808}\textbf{0.7456} &       & \emph{FoQuS}  & \cellcolor[rgb]{ .816,  .808,  .808}\textbf{0.6178} & \cellcolor[rgb]{ .816,  .808,  .808}\textbf{0.8014} & \cellcolor[rgb]{ .816,  .808,  .808}\textbf{0.7720} & \cellcolor[rgb]{ .816,  .808,  .808}\textbf{0.9162} & \cellcolor[rgb]{ .816,  .808,  .808}\textbf{0.9313} & \cellcolor[rgb]{ .816,  .808,  .808}\textbf{0.8077} \\
    \midrule
    \multirow{11}[2]{*}{CNN2D} & Uniform & 0.4116  & 0.6130  & 0.6850  & 0.7647  & 0.7861  & 0.6521  & \multirow{11}[2]{*}{CNN2D} & Uniform & 0.1855  & 0.4648  & 0.5143  & 0.5910  & 0.7054  & 0.4922  \\
          & Cal   & 0.3067  & 0.4717  & 0.5167  & 0.6671  & 0.7319  & 0.5388  &       & Cal   & 0.1818  & 0.3976  & 0.4794  & 0.5540  & 0.6409  & 0.4507  \\
          & CD    & 0.3450  & 0.4597  & 0.5262  & 0.6802  & 0.7021  & 0.5426  &       & CD    & 0.1737  & 0.4669  & 0.5031  & 0.5982  & 0.6843  & 0.4852  \\
          & Forgetting & 0.3544  & 0.4555  & 0.5886  & 0.6465  & 0.7022  & 0.5494  &       & Forgetting & 0.1692  & 0.4198  & 0.4777  & 0.5033  & 0.5920  & 0.4324  \\
          & GraNd & 0.3409  & 0.4305  & 0.4871  & 0.6542  & 0.6995  & 0.5224  &       & GraNd & 0.1814  & 0.4115  & 0.4487  & 0.5047  & 0.5584  & 0.4209  \\
          & Herding & 0.4343  & 0.6383  & 0.6821  & 0.7367  & 0.7786  & 0.6540  &       & Herding & 0.1857  & 0.4637  & 0.5274  & 0.6053  & 0.6706  & 0.4905  \\
          & k-Center Greedy & 0.3959  & 0.5709  & 0.6680  & 0.7112  & 0.7890  & 0.6270  &       & k-Center Greedy & 0.1866  & 0.4709  & 0.5540  & 0.5964  & 0.6860  & 0.4988  \\
          & Least Confidence & 0.3481  & 0.4730  & 0.5095  & 0.6440  & 0.6793  & 0.5308  &       & Least Confidence & 0.1807  & 0.3731  & 0.4512  & 0.4802  & 0.5716  & 0.4114  \\
          & Entropy & 0.3153  & 0.4419  & 0.5050  & 0.6234  & 0.6736  & 0.5118  &       & Entropy & 0.1710  & 0.3712  & 0.4579  & 0.5215  & 0.5797  & 0.4203  \\
          & Margin & 0.3606  & 0.4771  & 0.5133  & 0.6339  & 0.6868  & 0.5343  &       & Margin & 0.1752  & 0.3425  & 0.4775  & 0.5152  & 0.5943  & 0.4209  \\
          & \emph{FoQuS}  & \cellcolor[rgb]{ .816,  .808,  .808}\textbf{0.4515} & \cellcolor[rgb]{ .816,  .808,  .808}\textbf{0.6615} & \cellcolor[rgb]{ .816,  .808,  .808}\textbf{0.6929} & \cellcolor[rgb]{ .816,  .808,  .808}\textbf{0.7680} & \cellcolor[rgb]{ .816,  .808,  .808}\textbf{0.7903} & \cellcolor[rgb]{ .816,  .808,  .808}\textbf{0.6728} &       & \emph{FoQuS}  & \cellcolor[rgb]{ .816,  .808,  .808}\textbf{0.1900} & \cellcolor[rgb]{ .816,  .808,  .808}\textbf{0.5172} & \cellcolor[rgb]{ .816,  .808,  .808}\textbf{0.5764} & \cellcolor[rgb]{ .816,  .808,  .808}\textbf{0.6393} & \cellcolor[rgb]{ .816,  .808,  .808}\textbf{0.7380} & \cellcolor[rgb]{ .816,  .808,  .808}\textbf{0.5322} \\
    \bottomrule
    \end{tabular}}
  \label{tab:base}%
\end{table*}%

\section{Experiments}
\label{sec:Experiments}

\subsection{Experimental Setup}
In this paper, we conduct extensive experiments on three widely-used AMR datasets: RML2016.10a~\cite{o2016radio}, Sig2019-12~\cite{chen2021signet} and RML2018.01a~\cite{o2018over}. Unless otherwise specified, we uniformly use high SNR data for experiments.





We compare our method with $10$ existing coreset selection methods. Herein, we provide a detailed description of these methods as follows:
\textbf{Uniform} randomly extracts a certain proportion of samples from the original dataset with equal probability as the coreset, without relying on any data or model information. \textbf{Cal~\cite{margatina2021active}} aims to select contrasting samples, that is, samples with two different labels but similar feature representations. \textbf{CD~\cite{agarwal2020contextual}.} captures the diversity in spatial and semantic context of various object categories in a dataset and prioritizes those with high diversity. \textbf{Forgetting~\cite{toneva2018empirical}} counts the number of times a sample changes from a correct prediction to an incorrect prediction during training, giving priority to retaining samples that are easily forgotten. \textbf{GraNd~\cite{paul2021deep}} uses gradient information, such as the gradient norm, to measure the contribution of samples to model updates, giving priority to samples that have a greater impact on model parameters. \textbf{Herding~\cite{welling2009herding}} selects samples based on the distance between the coreset center and original dataset center in the feature space. It incrementally adds $1$ sample each time into the coreset that can minimize distance between two centers. \textbf{k-Center Greedy~\cite{sener2017active}} adopts a greedy strategy of the farthest point first to select the center point that maximizes the coverage radius in the representation space. \textbf{Least Confidence~\cite{coleman2019selection}} sorts the samples by the highest class probability (confidence) of the model, and prioritizes samples with the lowest model confident. \textbf{Entropy~\cite{coleman2019selection}} measures uncertainty based on the entropy of the predicted distribution and selects the sample with the highest entropy. \textbf{Margin~\cite{coleman2019selection}} calculates the difference between the probabilities of the top two categories. The smaller the difference, the closer the sample is to the decision boundary and needs to be selected first.

\subsection{Main Experiments}
\textbf{Comparison on RML2016.10a and Sig2019-12.} To evaluate the effectiveness of our method, we first conduct selection on RML2016.10a and Sig2019-12 and compare our method with the above-mentioned $10$ baselines. Specifically, for each method, $1\%, 5\%, 10\%, 20\% , 30\%$ of the original data are  sampled as coresets for evaluation. We use the same model structure for selection and testing. Here, we use CNN1D and CNN2D for experiments respectively. 
For a fair comparison, all coresets are class-balanced, i.e., the number of samples in each class is the same. We uniformly train the model for $50$ epochs using the coresets obtained by the baselines and our method for verification. In addition, to avoid randomness, each experiment is repeated $3$ times and the mean is reported. As shown in \cref{tab:base}, we find that as the sampling rate increases, the performance of all methods generally improves, which is because retaining more samples can provide more sufficient training. Besides, our proposed method significantly outperforms traditional baselines in most sampling rates, with the most significant improvement at extremely low sampling rates (1\%), while also maintaining a leading advantage at medium and high sampling rates.

\textbf{Comparison on RML2018.01a.} To further evaluate the effectiveness of our method, we then conduct selection on the more difficult RML2018.01a. All experimental settings are same as before. As shown in \cref{tab:RML2018.01a}. \emph{FoQuS} still outperforms all other baseline methods, demonstrating the effectiveness of our method.

\begin{table*}[htbp]
  \centering
  \caption{Comparison of our method against coreset selection baselines under various rates on RML2018.01a. The best results are marked in \textbf{boldface}. For each experimental result, we run $3$ times and report its mean.}
    \resizebox{0.99\textwidth}{!}{
    \begin{tabular}{cccccccc|cccccccc}
    \toprule
    \multirow{2}[4]{*}{Model} & \multirow{2}[4]{*}{Method} & \multicolumn{5}{c}{Rate}              & \multirow{2}[4]{*}{Avg} & \multirow{2}[4]{*}{Model} & \multirow{2}[4]{*}{Method} & \multicolumn{5}{c}{Rate}              & \multirow{2}[4]{*}{Avg} \\
\cmidrule{3-7}\cmidrule{11-15}          &       & 1\%   & 5\%   & 10\%  & 20\%  & 30\%  &       &       &       & 1\%   & 5\%   & 10\%  & 20\%  & 30\%  &  \\
    \midrule
    \multirow{11}[2]{*}{CNN1D} & Uniform & 0.4491  & 0.6280  & 0.7056  & 0.7617  & 0.7992  & 0.6687  & \multirow{11}[2]{*}{CNN2D} & Uniform & 0.4002  & 0.4756  & 0.5352  & 0.4756  & 0.5548  & 0.4883  \\
          & Cal   & 0.4123  & 0.5131  & 0.6387  & 0.7313  & 0.7948  & 0.6180  &       & Cal   & 0.3083  & 0.4806  & 0.5026  & 0.5396  & 0.5629  & 0.4788  \\
          & CD    & 0.4258  & 0.5070  & 0.7055  & 0.7615  & 0.8148  & 0.6429  &       & CD    & 0.3645  & 0.4833  & 0.5207  & \textbf{0.5685} & 0.5730  & 0.5020  \\
          & Forgetting & 0.4179  & 0.4983  & 0.6296  & 0.7208  & 0.7887  & 0.6111  &       & Forgetting & 0.3889  & 0.4681  & 0.5081  & 0.5398  & 0.5473  & 0.4904  \\
          & GraNd & 0.4450  & 0.6256  & 0.7019  & 0.7797  & 0.8033  & 0.6711  &       & GraNd & 0.4057  & 0.4895  & 0.5283  & 0.5009  & 0.5888  & 0.5026  \\
          & Herding & 0.4440  & 0.6162  & 0.7109  & 0.7831  & 0.7955  & 0.6699  &       & Herding & 0.4192  & 0.4805  & 0.5299  & 0.5520  & 0.5663  & 0.5096  \\
          & k-Center Greedy & 0.4351  & 0.6250  & 0.7147  & 0.7724  & 0.8139  & 0.6722  &       & k-Center Greedy & 0.4050  & 0.4895  & 0.5299  & 0.5580  & 0.5733  & 0.5111  \\
          & Least Confidence & 0.4191  & 0.5133  & 0.6376  & 0.7050  & 0.8000  & 0.6150  &       & Least Confidence & 0.3460  & 0.4831  & 0.5350  & 0.5549  & 0.5662  & 0.4970  \\
          & Entropy & 0.4236  & 0.5901  & 0.6596  & 0.7598  & 0.8164  & 0.6499  &       & Entropy & 0.3412  & 0.4872  & 0.5245  & 0.5642  & 0.5872  & 0.5009  \\
          & Margin & 0.4151  & 0.5571  & 0.6556  & 0.7848  & \textbf{0.8260} & 0.6477  &       & Margin & 0.3631  & 0.4807  & 0.5124  & 0.5457  & 0.5827  & 0.4969  \\
          & Ours  & \cellcolor[rgb]{ .816,  .808,  .808}\textbf{0.4497 } & \cellcolor[rgb]{ .816,  .808,  .808}\textbf{0.6321} & \cellcolor[rgb]{ .816,  .808,  .808}\textbf{0.7337} & \cellcolor[rgb]{ .816,  .808,  .808}\textbf{0.7901} & \cellcolor[rgb]{ .816,  .808,  .808}0.8250  & \cellcolor[rgb]{ .816,  .808,  .808}\textbf{0.6861} &       & Ours  & \cellcolor[rgb]{ .816,  .808,  .808}\textbf{0.4233} & \cellcolor[rgb]{ .816,  .808,  .808}\textbf{0.4908} & \cellcolor[rgb]{ .816,  .808,  .808}\textbf{0.5397} & \cellcolor[rgb]{ .816,  .808,  .808}0.5669  & \cellcolor[rgb]{ .816,  .808,  .808}\textbf{0.5966} & \cellcolor[rgb]{ .816,  .808,  .808}\textbf{0.5235} \\
    \bottomrule
    \end{tabular}}
  \label{tab:RML2018.01a}%
\end{table*}%

\textbf{Cross-architecture Generalization.}
In this paper, our method requires a model to quantify how “informative” each sample in the original training dataset is and selects the coreset accordingly. Therefore, to verify that our selection strategy is not dependent on a specific model structure, we further evaluate its performance on $2$ additional structures. In other words, we perform data selection using model A and then use model B to evaluate the performance of the selected subset, abbreviated as A$\rightarrow$B. As shown in \cref{tab:cross}, \emph{FoQuS} outperforms other baselines in most settings, demonstrating that our method has good generalization across model structures. Besides, we further evaluate our method using more advanced models SigNet~\cite{chen2021signet} and IQFormer~\cite{shao2024iqformer} on RML2016.10a. As shown in \cref{tab:signet}, \emph{FoQuS} also achieves good performance on more advanced models, further proving that our method can be generalized to more sophisticated model structures.

\begin{table*}[t]
  \centering
  \caption{Cross-architecture generalization of \emph{FoQuS}. The best results are marked in \textbf{boldface}. For each experimental result, we run $3$ times and report its mean. CNN1D$\rightarrow$CNN2D denotes using CNN1D for selection and CNN2D for evaluation.}
    \resizebox{0.99\textwidth}{!}{
    \begin{tabular}{c|cccccccc|cccccccc}
    \toprule
    \multirow{2}[4]{*}{Dataset} & \multirow{2}[4]{*}{Model} & \multirow{2}[4]{*}{Method} & \multicolumn{5}{c}{Rate}              & \multirow{2}[4]{*}{Avg} & \multirow{2}[4]{*}{Model} & \multirow{2}[4]{*}{Method} & \multicolumn{5}{c}{Rate}              & \multirow{2}[4]{*}{Avg} \\
\cmidrule{4-8}\cmidrule{12-16}          &       &       & 1\%   & 5\%   & 10\%  & 20\%  & 30\%  &       &       &       & 1\%   & 5\%   & 10\%  & 20\%  & 30\%  &  \\
    \midrule
    \multirow{22}[4]{*}{RML2016.10a} & \multirow{11}[2]{*}{CNN1D$\rightarrow$CNN2D} & Uniform & 0.4013  & 0.6449  & 0.6836  & 0.7558  & 0.7833  & 0.6538  & \multirow{11}[2]{*}{CNN1D$\rightarrow$LSTM} & Uniform & 0.1631  & \textbf{0.3160}  & 0.3130  & 0.5672  & 0.6561  & 0.4031  \\
          &       & Cal   & 0.3560  & 0.5251  & 0.6371  & 0.6948  & 0.7421  & 0.5910  &       & Cal   & 0.1699  & 0.2399  & 0.3250  & 0.2850  & 0.5795  & 0.3199  \\
          &       & CD    & 0.4230  & 0.6095  & 0.6822  & 0.7587  & 0.7792  & 0.6505  &       & CD    & 0.1751  & 0.2842  & 0.5252  & 0.6322  & 0.6670  & 0.4567  \\
          &       & Forgetting & 0.3953  & 0.5436  & 0.6423  & 0.7034  & 0.7530  & 0.6075  &       & Forgetting & 0.1687  & 0.2945  & 0.3417  & 0.6071  & 0.6236  & 0.4071  \\
          &       & GraNd & 0.4198  & 0.6114  & 0.6948  & 0.7579  & 0.7881  & 0.6544  &       & GraNd & 0.1765  & 0.2494  & 0.5339  & 0.6261  & 0.6535  & 0.4479  \\
          &       & Herding & 0.4197  & 0.6209  & 0.6643  & 0.7319  & 0.7573  & 0.6388  &       & Herding & 0.1586  & 0.2908  & 0.2781  & 0.6397  & 0.6644  & 0.4063  \\
          &       & k-Center Greedy & 0.3729  & 0.6181  & 0.6731  & 0.7590  & 0.7748  & 0.6396  &       & k-Center Greedy & 0.1538  & 0.1990  & 0.2741  & 0.6203  & 0.6417  & 0.3778  \\
          &       & Least Confidence & 0.3343  & 0.5050  & 0.6386  & 0.6811  & 0.7375  & 0.5793  &       & Least Confidence & 0.1773  & 0.2036  & 0.4684  & 0.2712  & 0.6324  & 0.3506  \\
          &       & Entropy & 0.2804  & 0.5095  & 0.6223  & 0.6769  & 0.7280  & 0.5634  &       & Entropy & 0.1753  & 0.2772  & 0.4674  & 0.5465  & 0.5728  & 0.4078  \\
          &       & Margin & 0.3056  & 0.5119  & 0.6286  & 0.6868  & 0.7237  & 0.5713  &       & Margin & 0.1721  & 0.2496  & 0.4617  & 0.5306  & 0.6422  & 0.4112  \\
          &       & \emph{FoQuS}  & \cellcolor[rgb]{ .816,  .808,  .808}\textbf{0.4317} & \cellcolor[rgb]{ .816,  .808,  .808}\textbf{0.6511} & \cellcolor[rgb]{ .816,  .808,  .808}\textbf{0.7101} & \cellcolor[rgb]{ .816,  .808,  .808}\textbf{0.7739} & \cellcolor[rgb]{ .816,  .808,  .808}\textbf{0.7892} & \cellcolor[rgb]{ .816,  .808,  .808}\textbf{0.6712} &       & \emph{FoQuS}  & \cellcolor[rgb]{ .816,  .808,  .808}\textbf{0.1864} & \cellcolor[rgb]{ .816,  .808,  .808}0.3041 & \cellcolor[rgb]{ .816,  .808,  .808}\textbf{0.5373} & \cellcolor[rgb]{ .816,  .808,  .808}\textbf{0.6577} & \cellcolor[rgb]{ .816,  .808,  .808}\textbf{0.7008} & \cellcolor[rgb]{ .816,  .808,  .808}\textbf{0.4773} \\
\cmidrule{2-17}          & \multirow{11}[2]{*}{CNN2D$\rightarrow$CNN1D} & Uniform & 0.4723  & 0.6243  & 0.7784  & 0.8192  & 0.8324  & 0.7053  & \multirow{11}[2]{*}{CNN2D$\rightarrow$LSTM} & Uniform & 0.1675  & 0.3001  & 0.2920  & 0.6054  & 0.6649  & 0.4060  \\
          &       & Cal   & 0.2889  & 0.5324  & 0.7205  & 0.8125  & 0.8251  & 0.6359  &       & Cal   & 0.1687  & 0.1972  & 0.2980  & 0.4584  & 0.5793  & 0.3403  \\
          &       & CD    & 0.3569  & 0.5496  & 0.7482  & 0.8225  & 0.8352  & 0.6625  &       & CD    & 0.1135  & 0.2238  & 0.2853  & 0.5001  & 0.6160  & 0.3477  \\
          &       & Forgetting & 0.3687  & 0.6192  & 0.7213  & 0.8037  & 0.8126  & 0.6651  &       & Forgetting & 0.1654  & 0.3368  & 0.3728  & 0.5279  & 0.6105  & 0.4027  \\
          &       & GraNd & 0.3274  & 0.5785  & 0.5863  & 0.7801  & 0.8058  & 0.6156  &       & GraNd & 0.1683  & 0.2280  & 0.3174  & 0.5369  & 0.6292  & 0.3760  \\
          &       & Herding & 0.4005  & 0.6490  & 0.7555  & 0.8237  & 0.8406  & 0.6939  &       & Herding & 0.1751  & 0.2513  & 0.4675  & 0.6045  & 0.6562  & 0.4309  \\
          &       & k-Center Greedy & 0.3512  & 0.6473  & 0.7397  & 0.7856  & 0.8358  & 0.6719  &       & k-Center Greedy & 0.1050  & 0.2244  & 0.3388  & 0.3675  & 0.6487  & 0.3369  \\
          &       & Least Confidence & 0.3511  & 0.4722  & 0.7180  & 0.7801  & 0.8042  & 0.6251  &       & Least Confidence & 0.1775  & 0.2325  & 0.4860  & 0.4910  & 0.6167  & 0.4007  \\
          &       & Entropy & 0.4186  & 0.6210  & 0.6817  & 0.8036  & 0.8000  & 0.6650  &       & Entropy & 0.1521  & 0.2825  & 0.4677  & 0.5226  & 0.6317  & 0.4113  \\
          &       & Margin & 0.4475  & 0.6190  & 0.6836  & 0.7829  & 0.8192  & 0.6704  &       & Margin & 0.1783  & 0.4494  & 0.5009  & 0.5659  & 0.6105  & 0.4610  \\
          &       & \emph{FoQuS}  & \cellcolor[rgb]{ .816,  .808,  .808}\textbf{0.4877} & \cellcolor[rgb]{ .816,  .808,  .808}\textbf{0.6895} & \cellcolor[rgb]{ .816,  .808,  .808}\textbf{0.7875} & \cellcolor[rgb]{ .816,  .808,  .808}\textbf{0.8313} & \cellcolor[rgb]{ .816,  .808,  .808}\textbf{0.8424} & \cellcolor[rgb]{ .816,  .808,  .808}\textbf{0.7277} &       & \emph{FoQuS}  & \cellcolor[rgb]{ .816,  .808,  .808}\textbf{0.1801} & \cellcolor[rgb]{ .816,  .808,  .808}\textbf{0.5032} & \cellcolor[rgb]{ .816,  .808,  .808}\textbf{0.5254} & \cellcolor[rgb]{ .816,  .808,  .808}\textbf{0.6367} & \cellcolor[rgb]{ .816,  .808,  .808}\textbf{0.6713} & \cellcolor[rgb]{ .816,  .808,  .808}\textbf{0.5033} \\
    \midrule
    \multirow{22}[4]{*}{Sig2019-12} & \multirow{11}[2]{*}{CNN1D$\rightarrow$CNN2D} & Uniform & 0.1863 & 0.4516 & 0.5523 & 0.6048 & 0.6835 & 0.4957  & \multirow{11}[2]{*}{CNN1D$\rightarrow$LSTM} & Uniform & 0.1090  & 0.4225  & 0.5335 & 0.7381 & 0.8582 & 0.5323  \\
          &       & Cal   & 0.1707 & 0.4521 & 0.5360  & 0.5712 & 0.6341 & 0.4728  &       & Cal   & 0.1057  & 0.3533 & 0.5236  & 0.6869 & 0.8855 & 0.5110  \\
          &       & CD    & 0.1745 & 0.4625 & 0.5069 & 0.5671 & 0.6419 & 0.4706  &       & CD    & 0.1026 & 0.3730  & 0.5107 & 0.6772 & 0.7636 & 0.4854  \\
          &       & Forgetting & 0.1776 & 0.4057 & 0.4901 & 0.5828 & 0.6812 & 0.4675  &       & Forgetting & 0.1118 & 0.3457 & 0.5115 & 0.6354 & 0.8283 & 0.4865  \\
          &       & GraNd & 0.1753 & 0.4007 & 0.4987 & 0.5750  & 0.6061 & 0.4512  &       & GraNd & 0.1104 & 0.3103 & 0.4563 & 0.6333  & 0.7352 & 0.4491  \\
          &       & Herding & 0.1808 & 0.4655 & 0.5196 & 0.6042 & 0.6857 & 0.4912  &       & Herding & 0.1099 & 0.4005 & 0.5262 & 0.7204 & 0.8832 & 0.5280  \\
          &       & k-Center Greedy & 0.1871 & 0.4599 & 0.5495 & 0.6101 & 0.6899 & 0.4993  &       & k-Center Greedy & 0.1129 & 0.4419 & \textbf{0.5592} & 0.7855 & 0.8962 & 0.5591  \\
          &       & Least Confidence & 0.1796 & 0.4556 & 0.5171 & 0.5749 & 0.6181 & 0.4691  &       & Least Confidence & 0.1141 & 0.2893 & 0.4297 & 0.5599 & 0.6952  & 0.4176  \\
          &       & Entropy & 0.1750  & 0.3917 & 0.5001 & 0.5438 & 0.6050  & 0.4431  &       & Entropy & 0.1115 & 0.3266 & 0.4430  & 0.6323 & 0.7363  & 0.4499  \\
          &       & Margin & 0.1777 & 0.4771 & 0.5010  & 0.5710  & 0.6204 & 0.4694  &       & Margin & 0.1085 & 0.3759 & 0.5160  & 0.6391 & 0.7103  & 0.4700  \\
          &       & \emph{FoQuS}  & \cellcolor[rgb]{ .816,  .808,  .808}\textbf{0.1938} & \cellcolor[rgb]{ .816,  .808,  .808}\textbf{0.4895} & \cellcolor[rgb]{ .816,  .808,  .808}\textbf{0.5577} & \cellcolor[rgb]{ .816,  .808,  .808}\textbf{0.6189} & \cellcolor[rgb]{ .816,  .808,  .808}\textbf{0.7296} & \cellcolor[rgb]{ .816,  .808,  .808}\textbf{0.5179} &       & \emph{FoQuS}  & \cellcolor[rgb]{ .816,  .808,  .808}\textbf{0.1180} & \cellcolor[rgb]{ .816,  .808,  .808}\textbf{0.4504} & \cellcolor[rgb]{ .816,  .808,  .808}0.5588 & \cellcolor[rgb]{ .816,  .808,  .808}\textbf{0.8251} & \cellcolor[rgb]{ .816,  .808,  .808}\textbf{0.9006} & \cellcolor[rgb]{ .816,  .808,  .808}\textbf{0.5706} \\
\cmidrule{2-17}          & \multirow{11}[2]{*}{CNN2D$\rightarrow$CNN1D} & Uniform & 0.5458 & 0.7318 & 0.7693 & 0.9183 & 0.9277 & 0.7786  & \multirow{11}[2]{*}{CNN2D$\rightarrow$LSTM} & Uniform & 0.1094  & 0.4112  & 0.5417 & 0.6977 & 0.7774 & 0.5075  \\
          &       & Cal   & 0.5199 & 0.7073 & 0.7516 & 0.9078 & 0.9282 & 0.7630  &       & Cal   & 0.1158 & 0.3420  & 0.5092  & 0.5913 & 0.8701 & 0.4857  \\
          &       & CD    & 0.5137 & 0.6722 & 0.7369 & 0.9169 & 0.9279 & 0.7535  &       & CD    & 0.1025  & 0.3829  & 0.5145 & 0.7025 & 0.8770  & 0.5159  \\
          &       & Forgetting & 0.5387  & 0.7247  & 0.7675 & 0.9192 & 0.9264  & 0.7753  &       & Forgetting & 0.1077 & 0.3990  & 0.5336 & 0.7344 & 0.8856  & 0.5321  \\
          &       & GraNd & 0.5027 & 0.6638 & 0.7389 & 0.7652  & 0.9306 & 0.7202  &       & GraNd & 0.1112 & 0.3911 & 0.4995 & 0.6415  & 0.8493 & 0.4985  \\
          &       & Herding & 0.5423 & 0.7405 & 0.7586 & 0.9110  & 0.9237 & 0.7752  &       & Herding & 0.1187 & 0.4143 & 0.5042 & 0.7887 & 0.8817 & 0.5415  \\
          &       & k-Center Greedy & 0.5586 & 0.7305 & 0.7656 & 0.9150  & 0.9225  & 0.7784  &       & k-Center Greedy & 0.1187 & \textbf{0.4283} & 0.5279 & 0.7483 & 0.8611 & 0.5369  \\
          &       & Least Confidence & 0.5447 & 0.6824 & 0.7504 & 0.7784 & 0.8873 & 0.7286  &       & Least Confidence & 0.1163 & 0.3835 & 0.5305 & 0.6995 & 0.8854 & 0.5230  \\
          &       & Entropy & 0.5213  & 0.6799 & 0.7568  & 0.9102  & 0.9296 & 0.7596  &       & Entropy & 0.1084 & 0.3322 & 0.5238 & 0.6847 & 0.7931 & 0.4884  \\
          &       & Margin & 0.5241 & 0.6927 & 0.7499 & 0.9115 & 0.7836 & 0.7324  &       & Margin & 0.1063 & 0.3869  & 0.5165 & 0.6466 & 0.8755 & 0.5064  \\
          &       & \emph{FoQuS}  & \cellcolor[rgb]{ .816,  .808,  .808}\textbf{0.6380} & \cellcolor[rgb]{ .816,  .808,  .808}\textbf{0.7460} & \cellcolor[rgb]{ .816,  .808,  .808}\textbf{0.8090} & \cellcolor[rgb]{ .816,  .808,  .808}\textbf{0.9213} & \cellcolor[rgb]{ .816,  .808,  .808}\textbf{0.9349} & \cellcolor[rgb]{ .816,  .808,  .808}\textbf{0.8098} &       & \emph{FoQuS}  & \cellcolor[rgb]{ .816,  .808,  .808}\textbf{0.1222} & \cellcolor[rgb]{ .816,  .808,  .808}0.4247 & \cellcolor[rgb]{ .816,  .808,  .808}\textbf{0.5517} & \cellcolor[rgb]{ .816,  .808,  .808}\textbf{0.8261} & \cellcolor[rgb]{ .816,  .808,  .808}\textbf{0.8934} & \cellcolor[rgb]{ .816,  .808,  .808}\textbf{0.5690} \\
    \bottomrule
    \end{tabular}}
  \label{tab:cross}%
\end{table*}%


\begin{table}[htbp]
  \centering
  \caption{Cross-architecture generalization of our method using more recent AMR models on RML2016.10a.}
  \resizebox{0.49\textwidth}{!}{
    \begin{tabular}{ccccccc}
    \toprule
    \multirow{2}[4]{*}{Model} & \multicolumn{5}{c}{Rate}              & \multirow{2}[4]{*}{Avg} \\
\cmidrule{2-6}          & 1\%   & 5\%   & 10\%  & 20\%  & 30\%  &  \\
    \midrule
    CNN1D$\rightarrow$SigNet & 0.6831  & 0.7543  & 0.8041  & 0.8343  & 0.8426  & 0.7837  \\
    CNN2D$\rightarrow$SigNet & 0.6876  & 0.7558  & 0.7970  & 0.8337  & 0.8443  & 0.7837  \\
    CNN1D$\rightarrow$IQFormer & 0.4602  & 0.7648  & 0.8239  & 0.8480  & 0.8799  & 0.7554  \\
    CNN2D$\rightarrow$IQFormer & 0.4375  & 0.7386  & 0.8330  & 0.8477  & 0.8858  & 0.7485  \\
    \bottomrule
    \end{tabular}}
  \label{tab:signet}%
\end{table}%

\subsection{Ablation Study}
\label{sec: Ablation Study}
In this paper, our \emph{FoQuS} score consists of three different metrics. To verify the contribution of each component, we carry out a comprehensive ablation study: we evaluate the performance of all single-metric, all pairwise combinations, and the full three-metric \emph{FoQuS} on RML2016.10a under 1\% sampling rate. We use the same CNN1D for selection and testing. As shown in \cref{tab:aba}. \emph{FoQuS} outperforms all other combinations, demonstrating the importance of metric diversification for the coreset.

\begin{table}[htbp]
  \centering
  \caption{Performance of \emph{FoQuS} with different metric combinations.}
  \resizebox{0.49\textwidth}{!}{
    \begin{tabular}{cccccccc}
    \toprule
    Forgetting Score & \ding{51}     & \ding{55}     & \ding{55}     & \ding{51}     & \ding{51}     & \ding{55}     & \ding{51} \\
    Persistent Error Score & \ding{55}     & \ding{51}     & \ding{55}     & \ding{51}     & \ding{55}     & \ding{51}     & \ding{51} \\
    Quality Score & \ding{55}     & \ding{55}     & \ding{51}     & \ding{55}     & \ding{51}     & \ding{51}     & \ding{51} \\
    \hline
    Acc   &   0.3564     &    0.5390    &   0.4703    &   0.4702     &   0.4932     &   0.5026     & \textbf{0.5410}  \\
    \bottomrule
    \end{tabular}}
  \label{tab:aba}%
\end{table}%

\section{Conclusion}
\label{sec:Conclusion}
This paper proposes a coreset selection method, \emph{FoQuS}, for AMR tasks. This method incorporates three complementary metrics: a forgetfulness score, a persistent error score and a quality score. The final \emph{FoQuS} score is obtained by normalizing these three metrics and adding them together. Next, we divide samples into three tiers based on the scores and extract a predetermined proportion of samples from each tier to form a diverse coreset. Finally, we demonstrate that the coreset obtained by \emph{FoQuS} outperforms existing image-based baselines on multiple AMR tasks.

\bibliographystyle{IEEEtran}
\bibliography{reference}

\vfill

\end{document}